\documentclass[final,3p,times]{elsarticle} 
\usepackage[utf8]{inputenc}
\usepackage[draft]{changes}
\usepackage[inline]{enumitem}
\definechangesauthor[color=blue]{PC}
\usepackage{lineno,hyperref}
\usepackage{amsmath,amssymb,amsfonts}
\usepackage{algorithmicx}
\usepackage{booktabs}
\usepackage{graphicx}
\usepackage{caption}
\usepackage{subfigure}
\usepackage[font=small,labelfont=bf, figurename=Fig.]{caption} 

\usepackage{textcomp}
\usepackage{todonotes}
\usepackage{tabularx}
\usepackage{gensymb}
\usepackage[utf8]{inputenc}
\usepackage{mathtools, nccmath}
\usepackage{subcaption}
\usepackage{booktabs}
\usepackage{cleveref}
\usepackage{chemformula}
\usepackage[version=4]{mhchem}
\usepackage{booktabs, multirow} 
\usepackage{soul}
\usepackage{tablefootnote}
\usepackage{setspace}
\usepackage{comment}
\usepackage[inline]{enumitem}  
\usepackage{algorithm}
\usepackage{algpseudocode}
\usepackage{makecell}

\modulolinenumbers[1]
\biboptions{authoryear}
\definecolor{DarkGreen}{rgb}{0.2,0.5,0.2}
\hypersetup{
    colorlinks=true, 
    citecolor=DarkGreen,  
    linkcolor=black, 
    filecolor=black,
    urlcolor=blue  
}
 
\makeatletter
\newcommand{\inlineitem}[1][]{%
\ifnum\enit@type=\tw@
    {\descriptionlabel{#1}}
  \hspace{\labelsep}%
\else
  \ifnum\enit@type=\z@
       \refstepcounter{\@listctr}\fi
    \quad\@itemlabel
    \hspace{\labelsep}%
\fi}
\makeatother
\begin{document}
\begin{frontmatter}

\title{Macroscopic Characteristics of Mixed Traffic Flow with Deep Reinforcement Learning Based Automated and Human-Driven Vehicles}

\author[mymainaddress]{Pankaj Kumar}
\author[mymainaddress]{Pranamesh Chakraborty\corref{mycorrespondingauthor}}
\author[mymainaddress1]{Subrahmanya Swamy Peruru}

\cortext[mycorrespondingauthor]{Corresponding author\\
Pranamesh Chakraborty: pranames@iitk.ac.in,  (+91-512-259-2146), Pankaj Kumar: pankajkmr22@iitk.ac.in, Subrahmanya Swamy Peruru: swamyp@iitk.ac.in, (+91-512-679-2189)}

\address[mymainaddress]{Department of Civil Engineering, Indian Institute of Technology Kanpur, Kanpur-208016, U.P., India}
\address [mymainaddress1]{Department of Electrical Engineering Indian Institute of Technology Kanpur, Kanpur-208016, U.P., India}

\begin{abstract}
Automated Vehicle (AV) control in mixed traffic, where AVs coexist with human-driven vehicles, poses significant challenges in balancing safety, efficiency, comfort, fuel efficiency, and compliance with traffic rules while capturing heterogeneous driver behavior. Traditional car-following models, such as the Intelligent Driver Model (IDM), often struggle to generalize across diverse traffic scenarios and typically do not account for fuel efficiency, motivating the use of learning-based approaches. Although Deep Reinforcement Learning (DRL) has shown strong microscopic performance in car-following conditions, its macroscopic traffic flow characteristics remain underexplored.
This study focuses on analyzing the macroscopic traffic flow characteristics and fuel efficiency of DRL-based models in mixed traffic. A Twin Delayed Deep Deterministic Policy Gradient (TD3) algorithm is implemented for AVs' control and trained using the NGSIM highway dataset, enabling realistic interaction with human-driven vehicles. Traffic performance is evaluated using the Fundamental Diagram (FD) under varying driver heterogeneity, heterogeneous time-gap penetration levels, and different shares of RL-controlled vehicles. A macroscopic level comparison of fuel efficiency between the RL-based AV model and the IDM is also conducted.
Results show that traffic performance is sensitive to the distribution of safe time gaps and the proportion of RL vehicles. Transitioning from fully human-driven to fully RL-controlled traffic can increase road capacity by approximately 7.52\%. Further, RL-based AVs also improve average fuel efficiency by about 28.98\% at higher speeds (above 50 $km/h$), and by 1.86\% at lower speeds (below 50 $km/h$) compared to the IDM. Overall, the DRL framework enhances traffic capacity and fuel efficiency without compromising safety.
\end{abstract}


\end{frontmatter}

\section{Introduction}

\label{sec: introduction}
Advancements in autonomous driving technologies offer significant promise for enhancing road safety, traffic efficiency, and environmental sustainability by minimising human error and optimising vehicular control strategies \citep{hart2024towards}. However, such environments introduce behavioral heterogeneity, which strongly influences both microscopic vehicle interactions and macroscopic traffic flow characteristics, including speed, density, and flow \citep{chen2022fundamental, yao2022fundamental,  wang2023fundamental, jiang2024fundamental, makridis2025fundamental}.
A series of studies have been conducted to explore the potential of automated vehicles (AVs) in future mixed traffic environments coexisting with human-driven vehicles \citep{zou2018impact, qin2024energy, agarwal2024dynamic}. Building on this foundation, studies on connected vehicles have shown that Adaptive Cruise Control (ACC) systems and Cooperative Adaptive Cruise Control (CACC) improve responsiveness, driving comfort, and overall traffic flow stability \citep{takahama2018model,chen2021mixed,qin2021lighthill,li2023fundamental,wang2023fundamental}. 
The existing ACC and CACC strategies typically rely on approaches such as consensus-based optimization \citep{jia2016platoon, wang2017capturing, salvi2017design}, Model Predictive Control (MPC) \citep{takahama2018model, chen2021mixed, zhai2022ecological}, dynamic programming \citep{ma2021eco, li2022cooperative}, and other control-based methods\citep{zhai2022ecological}. However, in recent years, Deep Reinforcement Learning (DRL) has emerged as a promising alternative to traditional vehicle control approaches. DRL-based strategies have demonstrated the capability of learning complex driving behaviors, adapting to diverse traffic conditions, and achieving better fuel efficiency than human drivers \citep{cheng2023online, yang2024eco, kumar2025deep}. 
The existing DRL-based research has primarily focused on microscopic driving behavior, particularly car-following control, where agents learn from interactions and generalize across diverse and uncertain traffic conditions \citep{zhu2020safe, gao2021autonomous, yang2024eco, hart2024towards, marin2025follow, kumar2025deep}. However, there remains a lack of a systematic investigation of the macroscopic traffic flow implications of deploying DRL-controlled automated vehicles.

At the macroscopic level, traffic behavior is classically captured by the Fundamental Diagram (FD), which describes the empirical relationship between flow, density, and speed. As one of the most foundational concepts in traffic flow theory and transportation engineering applications, the FD provides critical insights into roadway capacity, traffic stability, and the efficiency of a transportation network. Considerable research has been conducted on modelling the properties of the FD and developing robust methods for its estimation from simulated and real-world data \citep{he2015simple, shi2021constructing,  chen2022fundamental, yao2022fundamental, wang2023fundamental, gloudemans202324, jiang2024fundamental, chaudhari2025mitra, makridis2025fundamental}.
However, the existing research on mixed traffic flow has focused on investigating the macroscopic behavior, using rule-based controllers or model-driven strategies embedded within ACC and CACC frameworks, analyzing how these technologies influence aggregate traffic characteristics such as flow, density, and speed \citep{chen2022fundamental,yao2022fundamental, wang2023fundamental, jiang2024fundamental, makridis2025fundamental}.

Early investigations into the macroscopic impact of AVs primarily focused on traffic flow capacity, often employing simulation-based methods to examine the effects of varying AV market penetration rates and platooning intensity \citep{liu2018modeling, ghiasi2019mixed, hu2021analytical, wang2023fundamental}. For instance, previous studies such as \citep{yao2022fundamental, li2023fundamental, wang2023fundamental} used microscopic simulations to project significant potential gains in road capacity. However, a key limitation of these pioneering studies was their reliance on theoretical models, which often predicted substantial capacity improvements under optimistic assumptions such as instantaneous vehicle response and very short headways \citep{zhou2020modeling}. However, subsequent empirical investigations have demonstrated that such assumptions may not hold in practice. Field experiments with ACC systems, such as those conducted by \citep{shi2021constructing}, showed that while the overall triangular structure of the FD remains valid, achievable road capacity is highly sensitive to headway settings. Research on connected vehicles further demonstrates that ACC and CACC can enhance traffic flow while preserving the general FD structure; however, the magnitude of these benefits depends not only on penetration rates but also on platoon formation and the extent of vehicle-to-vehicle communication  \citep{wang2023fundamental,li2022deep, jiang2024fundamental, makridis2025fundamental}.

While ACC and CACC represent structured, rule-based automation, learning based longitudinal controllers introduce greater behavioral flexibility and can optimise multiple objectives simultaneously, including safety, efficiency, comfort, and energy consumption. Unlike classical analytical car-following models such as the IDM \citep{treiber2013traffic}, which in this study represents human-driven behavior, these controllers are not constrained by predefined functional forms. A recent study \citep{kumar2025deep} shows that such approaches can substantially modify microscopic driving behavior, particularly acceleration patterns and time-gap regulation, thereby affecting macroscopic traffic flow characteristics.

Importantly, RL-based car-following models are not trained to imitate human driving behavior; instead, they learn policies directly from interaction and reward signals. 
Therefore, it is extremely important to understand how it will behave at the macroscopic level beyond the microscopic improvements typically reported \citep{zhu2020safe, okhrin2022simulating, hart2024towards, yang2024eco, marin2025follow, ben2025reinforcement, kumar2025deep}. Differences between human drivers and DRL-controlled vehicles may introduce pronounced behavioral heterogeneity, which can affect roadway capacity and utilisation. Moreover, beyond traffic flow efficiency, longitudinal control strategies directly affect vehicle energy consumption, as fuel usage is highly sensitive to acceleration behavior \citep{cheng2023online, kumar2025deep}. 

Motivated by these gaps, this study investigates macroscopic traffic flow dynamics arising from interactions between DRL-based AVs and human-driven vehicles represented by the IDM. By leveraging empirically grounded trajectory data and systematically varying safe time gaps and vehicle penetration rates, the study provides insight into how learning-based longitudinal control alters fundamental diagram characteristics and traffic performance. Although DRL-based driving strategies often incorporate fuel or energy efficiency into their optimization objectives, their implications have rarely been examined at the macroscopic level. Accordingly, this study also includes a macroscopic fuel-efficiency analysis, enabling a joint assessment of traffic efficiency and energy performance in mixed traffic conditions.

Therefore, the main contributions of this study are summarized as follows: 
\begin{itemize}
 \item Developing FD for RL-based AVs to analyse their macroscopic traffic flow characteristics.
 
 \item Study the impact of RL-based driver heterogeneity (in terms of different aggressiveness quality of drivers) modelled through different safe time gap settings on the macroscopic traffic flow characteristics using the FD framework.
 
 \item Construct FDs for a mixed traffic environment consisting of RL-based AVs and human-driven vehicles (modelled using an IDM-based car following model).
 
 \item Evaluate the fuel efficiency of the RL-based car following model by comparing with IDM to understand how learning-based (RL-model) control strategies influence energy performance at the macroscopic level.
 
\end{itemize}

To achieve these objectives, sequential vehicle trajectories are generated using a DRL-based longitudinal control framework and the IDM, enabling a unified microscopic–macroscopic analysis of traffic flow and energy performance.

Apart from the introduction and objective description provided in Section~\ref{sec: introduction}, the remainder of the paper is organised as follows. Section~\ref{sec: methodology} outlines the proposed methodology, including the formulation of the state space, action set, reward function, RL algorithm and the estimation of macroscopic traffic flow variables used in this study. Section~\ref{sec: data} describes the data description and environment setup. Section~\ref{sec: test_results} presents the RL training and experimental results along with a comparative analysis across different driving scenarios. Finally, Section~\ref{sec: conclusion} summarises the key findings and contributions of the work and highlights potential directions for future research.

\section{Design Methodology}
\label{sec: methodology}

The primary objective of this study is to understand the macroscopic traffic behavior of DRL-based autonomous agents. Therefore, the methodology primarily involves selecting an appropriate RL-based car-following model and estimating macroscopic traffic variables in a traffic stream to construct the FD. The DRL-based car-following model adopted in this study is based on our previous work \citep{kumar2025deep}. 
We first discuss the key elements required to construct the RL model, including the state representation and action space (Section~\ref{subsec: state_action}). The reward function is formulated in Section~\ref {subsec: reward_formulation}, followed by the implementation details of the RL algorithm in Section~\ref{sec: RL_algorithm}. Finally, the methodology for estimating macroscopic traffic flow variables is then outlined in Section~\ref{sec: macroscopic}.

\subsection{Action space and state dynamics}
\label{subsec: state_action}
In this study, the action space is defined as the RL agent's (ego vehicle's) longitudinal acceleration/deceleration, represented by the continuous range of feasible acceleration values: $a^{(t)} \in [a_{min}, a_{max}]$, where $a_{min}$ = -4 $m/s^2$ and $a_{max}= 2 \ m/s^2$. The upper bound $a_{max}$ corresponds to a comfortable acceleration level during normal driving conditions, such as when the leading vehicle is farther ahead, enabling the follower vehicle to accelerate smoothly. On the other hand, the lower bound $a_{min}$ on the maximum deceleration the agent can apply, which reflects emergency braking scenarios under dry road conditions \citep{treiber2015comparing}. Please note that the RL agent's action variable (i.e., acceleration) is similar to that used in a typical car-following model, such as the IDM \citep{treiber2013traffic}.

The state variables are critical for enabling the agent to learn the optimal policy, and they are defined as the set of observable variables that directly influence the agent’s actions at each time step. The state vector include following components: (i) speed of ego vehicle $(v_n^{(t)})$, which helps regulate comfort and efficiency; (ii) relative velocity between the leading vehicle and the ego vehicle $\left(\Delta v_{n-1, n}^{(t))}\right)$ captures the urgency of responding to changes in the leader’s motion; (iii) bumper to bumper gap between leader and follower vehicle $\left(S_{n-1, n}^{(t)}\right)$ ensures collision avoidance and compliance with traffic safety norms. (iv) Safe time gap $(T)$ of the ego vehicle, which represents the temporal gap between the ego and the leading vehicle. This measure not only promotes safe car-following behavior but also enables the RL agent to emulate a range of driving styles, from aggressive (shorter time gaps) to conservative (longer time gaps). Here, the subscripts $n-1$ and $n$ denote the leader and ego vehicles, respectively. The state space at each time step $t$ is formally represented as in Eq.~\eqref{eq: state}.

\begin{equation}
    s^{(t)} = \left(v_n^{(t)}, \Delta v_{n-1, n}^{(t)}, S_{n-1, n}^{(t)}, T \right)
    \label{eq: state}
\end{equation}
The state transitions are computed using vehicle kinematics, and the agent's actions control the evolution of speed, relative velocity, and spacing. The state transition equations are shown in Eq.~\eqref{eq: state_update}.

\begin{equation}
\begin{aligned}
v_n^{(t + 1)} &= v_n ^{(t)} + a_n^{(t)} \times \Delta T \\
\Delta v_{n-1,n}^{(t + 1)} &= v_{n-1}^{(t + 1)} - v_n^{(t + 1)} \\
S_{n-1,n}^{(t + 1)} &= S_{n-1,n}^{(t)} + \Delta v_{n-1,n}^{(t)} \times \Delta T \\
\end{aligned}
\label{eq: state_update}
\end{equation}

Here, $a_n^{(t)}$ denotes the action selected by the agent at time $t$, $\Delta T$ refers to the time interval between two consecutive simulation time steps. These equations describe how the ego vehicle’s motion and interaction with the leading vehicle evolve in response to the agent’s control actions.

\subsection{Reward formulation}
\label{subsec: reward_formulation}

The reward function is adopted from our previous work \citep{kumar2025deep}, originally developed for longitudinal control at signalized intersections. In this study, it is adapted for highway traffic by removing the signal-related components. The function guides the RL agent’s learning and long-term decision-making through multiple components addressing safety (collision avoidance), traffic efficiency, comfort, and compliance with speed limits. It also promotes adaptation to different driving styles (e.g., aggressive or conservative) by dynamically adjusting the safe time gap based on traffic conditions and leader behavior. The following section details the corresponding reward components.

\subsubsection*{Safety and driving comfort}
Our primary objective is to ensure the passenger's safety and comfort while the vehicle is in motion or following the leader vehicle. Accordingly, the RL agent must avoid rear-end collisions with the lead vehicle, regardless of the lead vehicle’s behavior. To capture safety and comfort, two widely recognized indicators are used: Time to Collision (TTC) and jerk. TTC is a commonly used metric for safety assessment, while jerk (the rate of change of acceleration) is often associated with passenger comfort \cite{zhu2020safe, yang2024eco}. TTC quantifies the time remaining before a potential collision would occur if both the ego and lead vehicles continue at their current speeds. It is calculated as the ratio of relative distance to relative speed between the leader and follower vehicles. The safety-related reward based on TTC is defined in Eq.~\eqref{eq: f_TTC}. When TTC exceeds the threshold, no penalty is applied. To further discourage collisions, the most severe safety violation, a large fixed penalty (-50) is applied to any collision.

\begin{equation}
f_{\text{TTC}}^{(t)} = \begin{cases} \left( \frac{TTC ^{(t)}}{TTC_{\text{threshold}}} \right)^2 - 1 & \text{if } 0 \leq TTC^{(t)} < TTC_{\text{threshold}} \\0 & \text{otherwise}
\end{cases}
\label{eq: f_TTC}
\end{equation}

For driving comfort, we have used a reward term that penalizes high jerk values, as shown in Eq.~\eqref{eq: f_jerk}. Reducing jerk encourages smoother acceleration profiles, improving passenger comfort. Similar comfort-based reward formulations have been adopted in previous studies \citep{zhou2019development, yang2024eco, kumar2025deep}.

\begin{equation}
f_{\text{Jerk}}^{(t)} = -\left[\frac{|j^{(t)}|}{\frac{a_{\text{max}} - a_{\text{min}}}{\Delta T}}\right]^{\frac{1}{4}}
\label{eq: f_jerk}
\end{equation}
Here, $j$ represents the jerk, while $a_{min}$ and $a_{max}$ correspond to the minimum $(- 4 \ m/s^2)$ and maximum $(2\ m/s^2)$ allowable acceleration values, respectively. $\Delta T$ denotes the time step. In Eq.~\eqref{eq: f_jerk}, the term $\frac{a_{\text{max}} - a_{\text{min}}}{\Delta T}$ serves to normalise the absolute value in the interval [0,1] of the jerk-based reward component.

\subsubsection*{Driving efficiency}
Apart from the TTC reward, which penalizes the RL agent for keeping small TTC values, leading to a larger safety margin, such behavior can negatively impact traffic efficiency by reducing roadway capacity. Therefore, to balance safety and efficient traffic flow, an additional efficiency-related reward based on space headway is incorporated. This space headway-based reward is adopted from our previous work \citep{kumar2025deep}. The traffic efficiency reward formulation is presented in Eq.~\eqref{eq: f_eff}.

\begin{equation}
    f_{Eff}^{(t)} = \frac{f_{\text{LN}(S^{(t)}|\mu,\sigma)}}{f_{\text{LN}(S^{*(t)}|\mu,\sigma)}}; f_{\text{LN}(S^{*(t)}|\mu,\sigma)}>0
    \label{eq: f_eff}
\end{equation}
where, 
\begin{equation}
f_{\text{LN}(S^{(t)}|\mu,\sigma)}^{(t)} = \frac{1}{S^{(t)} \sigma \sqrt{2\mu}} \exp\left(-\frac{(\ln(S^{(t)}) - \mu^2}{2\sigma^2}\right); S^{(t)}>0
\label{eq: lognorm}
\end{equation}
with $\mu = ln(S^{*(t)}) + \sigma^2$, $\sigma = 1$. 

\begin{equation}
    S^{*(t)}= S_0+v_n^{(t)}T
    \label{eq: space_headway}
\end{equation}

Here, $f_{\text{LN}(S^{(t)}|\mu,\sigma)}^{(t)}$ denotes the probability density function of the lognormal distribution characterised by the parameters mean $(\mu)$ and standard deviation $(\sigma)$. The term $S^{*(t)}$ refers to the desired space headway, which is calculated using Eq.~\eqref{eq: space_headway} where $S_0$ represent the minimum bumper-to-bumper distance when vehicles are stationary, $T$ denotes the safe time gap, and $v_n^{(t)}$ denotes the current speed of the ego vehicle. The RL agent receives the maximum reward when the actual space headway $(S^{(t)})$ is equal to the desired space headway $(S^{*(t)})$, and decreases as the deviation increases.
Here, the safe time-gap $(T)$ parameter allows the reward function to capture \textit{\textbf{driver heterogeneity}}. By varying $T$, the RL agent's behavior can mimic different types of drivers. A lower value of $T$ results in shorter following distances, mimicking a more aggressive or closely following driving behavior. Conversely, a higher $T$ value leads to larger headway, indicating a more conservative driving style in which the vehicle maintains a greater buffer from the leader. The effect of different variations of $T$ on the agent's behavior and resulting trajectories is discussed in detail in Section ~\ref{sec: driver_heterogeneity}. 

\subsubsection*{Adherence to desired speed}
Beyond ensuring safety, efficiency, and comfort, the RL agent must maintain the desired speed without exceeding the speed limit when the LV is far away. To penalize overspeeding, a reward component based on deviation from the speed limit is introduced (Eq.~\eqref{eq: f_speed}). A quadratic penalty structure is adopted so that the further the agent's speed exceeds the speed limit, the greater the penalty it incurs. The denominator in Eq.~\eqref{eq: f_speed} acts as a normalisation factor to appropriately scale the penalty magnitude.

\begin{equation}
f_{\text{Speed}}^{(t)} = \begin{cases} - \left(\frac{v_n ^{(t)} - v_{\text{limit}}}{v_{\text{limit}}}\right)^2 & \text{if } v_n^{(t)} > v_{\text{limit}} \\
0 & \text{otherwise}
\end{cases}
\label{eq: f_speed}
\end{equation}

\subsubsection*{Fuel efficiency}

Along with the safety, efficiency, comfort, and adherence to traffic rules, the RL agent (ego vehicle) is also designed to optimise fuel efficiency. Fuel efficiency here refers to the distance travelled per unit of fuel consumed. To promote energy-efficient driving behavior, a dedicated reward function has been used that assigns higher scores to actions resulting in greater distance covered per unit of fuel. A similar reward function was also used in eco-driving studies \citep{yang2024eco}. At each time step, the fuel-efficiency reward $f_{Fuel}^{(t)}$ is calculated using Eq.~\eqref{eq: fuel_reward}.

\begin{equation}
    f_{Fuel}^{(t)} = \left[\log_m\left(\frac{D^{(t)}}{r_{Fuel}^{(t)}}+1\right)\right]^5
    \label{eq: fuel_reward}
\end{equation}

Here, $D^{(t)}$ is the travel distance at time step $t$, while $r_{Fuel}^{(t)}$ represents the fuel rate at each time step. The parameter $m$ acts as a scaling factor. employed to compress the wide range of possible efficiency values, while the exponential term enhances the sensitivity of the measure, particularly when fuel efficiency is low. To compute instantaneous fuel use, the Virginia Tech Comprehensive Power-based Fuel Model, Type 1 (VT-CPFM-1) \citep{rakha2011virginia}, is applied. This model captures the relationship between vehicle power demand and fuel consumption, making it suitable for estimating fuel usage under various driving conditions.

The instantaneous fuel consumption rate at time $t$, $r_{Fuel}^{(t)}$ is calculated as shown in Eq.~\eqref{eq: fuel_rate} \citep{rakha2011virginia} 
\begin{equation}
    r_{Fuel}^{(t)} = 0.000341 + 0.0000583 \times P^{(t)} + 0.000001 \times (P^{(t)})^2
    \label{eq: fuel_rate}
\end{equation}

Here, $P^{(t)}$ represents the power exerted by the vehicle driveline (in kW) at time $t$. This driveline power is computed as shown in Eq.~\eqref{eq: fuel_power} \citep{rakha2011virginia}
\begin{equation}
     P^{(t)} = max\left(0, \frac{(R^{(t)}+1.04 \times m \times a^{(t)})}{3600 \eta_d} \times v^{(t)} \right)
     \label{eq: fuel_power}
\end{equation}
with $m$ representing the vehicle mass (2000 kg), $a^{(t)}$ the vehicle acceleration at time 
$t$, $\eta_d$ the driveline efficiency, and $v^{(t)}$ the vehicle speed at time $t$ (m/s). The power $ P^{(t)}$ depends on the total resistance force $ R^{(t)}$, which in turn accounts for aerodynamic drag, rolling resistance, and road grade, and is given by \citep{rakha2011virginia}
\begin{equation}
     R^{(t)} = \frac{\rho}{25.92} C_D C_h A_f (v^{(t)})^2 + g\times m \frac{C_r}{1000} (c_1 v^{(t)} +c_2) + g \times m\times G^{(t)}
     \label{eq: fuel_resistance}
\end{equation}
Here $\rho$ is the air density sea level at $15^{\circ C}$ (equal to 1.225 $kg/m^3$), $C_D$ is the vehicle drag coefficient equal to 0.3, $C_h$ is the altitude correction factor equal to 0.85, $A_f$ is the frontal area of the vehicle equal to 2.015 $m^2$, $g$ is gravitational acceleration (9.806 $m/s^2$), $C_r$ is the rolling resistance conversion factor (1.750), $c_1$ and $c_2$ are the rolling resistance parameter with values 0.0328 and 4.575 respectively, and $G^{(t)}$ represents the road gradient at time t with value 0.

\subsubsection*{Total reward}
\label{sec; total_reward}
At each simulation time step, the total reward is computed as a weighted sum of the individual reward components. These components collectively address key driving objectives such as safety, efficiency, comfort, adherence to traffic rules, and fuel efficiency. The reward function is mathematically represented as:

\begin{equation}
  R^{(t)} = 
  \begin{cases} 
    w_1 f_{\text{TTC}}^{(t)} + w_2 f_{\text{Eff}}^{(t)} + w_3 f_{\text{Jerk}}^{(t)} + w_4 f_{\text{Speed}}^{(t)} +  w_5 f_{\text{Fuel}}^{(t)} & \text{if } S > 0 \\
    w_6 \times {\text{Collision penalty}} & \text{otherwise}.
  \end{cases}
  \label{eq: reward}
\end{equation}

Here, $w_1$, $w_2$, $w_3$, $w_4$, $w_5$, and $w_6$ represent the weighting factors associated with the individual reward components, determining their relative influence in the overall reward formulation. All parameter values are adopted from our previous work \citep{kumar2025deep}. 

\subsection{Reinforcement learning (RL) algorithm description}
\label{sec: RL_algorithm}
The primary objective of RL is to develop a strategy (called a policy) that enables an agent to interact with the environment to maximise cumulative rewards over time \citep{sutton2018reinforcement}. In this study, the environment consists of a two-vehicle car-following scenario in which the follower vehicle is controlled by the RL agent. The interaction between the agent and environment is typically modelled as a discrete-time Markov Decision Process (MDP). At each time $t$, the RL agent observes the state of the environment $(s^{(t)})$, which influenced its decision. Based on this state, the agent selects an action $(a^{(t)})$ according to policy $\pi(a^{(t)}, s^{(t)})$. In response, the environment produces a reward  $(r^{(t)})$, providing feedback to the agent, and transitions to a new state $(s^{(t+1)})$ following the transition dynamics $p(s^{(t+1)}|(a^{(t)})$. The objective is to determine the optimal policy $(\pi^*)$ that maximize the expected discounted cumulative rewards $E_\pi\left[\sum_{i= 0}^\infty \gamma^{i}r_{(t+i+1)}\right]$, where $\gamma \in (0,1]$ is the discount factor. The definitions of the state, action, and reward are provided in Sections~\ref{subsec: state_action} and ~\ref{subsec: reward_formulation}. 

Several car-following studies have adopted the Deep Deterministic Policy Gradient (DDPG) algorithm \citep{lillicrap2015continuous} to learn optimal control policies \citep{huang2018car,zhou2020modeling, liu2018modeling,hart2024towards,yang2024eco,kumar2025deep}. Despite its popularity, DDPG suffers from limitations, including sensitivity to hyperparameters, overestimation bias in Q-value estimates, and general instability during training. To address these limitations, this study employs the Twin Delayed Deep Deterministic Policy Gradient (TD3) algorithm \citep{fujimoto2018addressing}, which has demonstrated superior and more stable performance in car-following applications \citep{kumar2025deep}. TD3 is an off-policy, model-free algorithm that utilises twin critic networks to reduce overestimation bias, delayed policy updates for improved stability, and target policy smoothing to prevent exploitation of sharp value function peaks. As the focus of this work is on analyzing macroscopic traffic flow characteristics under mixed traffic conditions rather than algorithmic comparison, TD3 is adopted exclusively. Further details regarding the implementation and hyperparameter settings of the RL model are available in our previous study \citep{kumar2025deep}.

\subsection{Estimation of macroscopic traffic variables}
\label{sec: macroscopic}
Typically, traffic flow characteristics such as density, flow, and speed can be quantified using two distinct approaches: microscopic and macroscopic methods. 
\subsubsection*{Microscopic method}
In this method, the density is directly estimated from the average spacing (S) under the equilibrium conditions, as expressed in Eq.~\eqref{eq: micro_relation}. The speed is obtained by assuming steady-state equilibrium, which implies that the speed of all vehicles is identical and equal to the macroscopic speed $(v_e(s))$. The traffic flow is then calculated using the fundamental equilibrium relation shown in Eq.~\eqref{eq: micro_relation}.
\begin{equation}
    k = \frac{1000}{S},  \quad 
    v = v_e(s),  \quad
    q = k \times v_e(s)
    \label{eq: micro_relation}
\end{equation}
Previous studies have utilised the above relation to calculate macroscopic traffic flow variables \citep{shi2021constructing, chen2022fundamental, wang2023fundamental, jiang2024fundamental}. However, this approach is not suitable because the assumption of steady-state equilibrium rarely holds in practice, vehicle speeds are heterogeneous, spacings vary dynamically, and stochastic effects lead to fluctuations that cannot be captured by equilibrium-based formulations. As a result, directly applying these equations may oversimplify the dynamics and fail to reflect realistic traffic conditions.

\subsubsection*{Macroscopic method}
\begin{figure}[h]
    \centering
    \includegraphics[width=\linewidth]{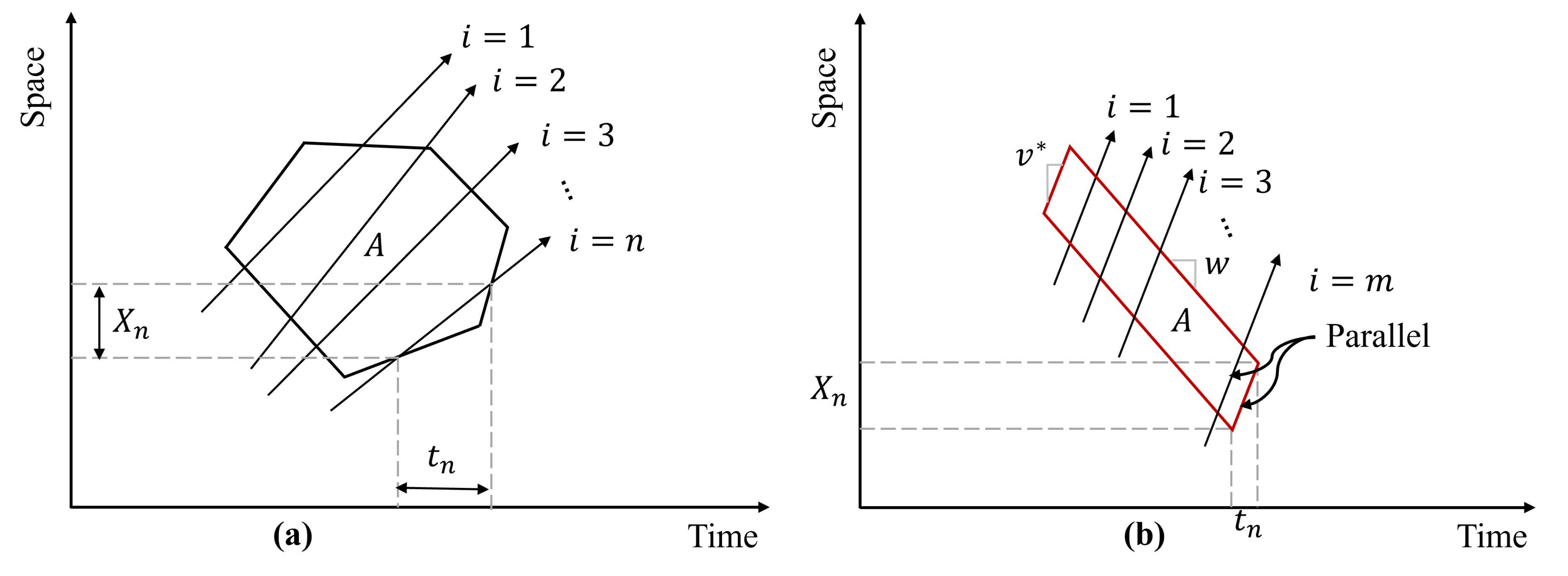}
    \caption{General schematic representation of (a) time-space region to measure traffic flow characteristics, and (b) parallelogram region with trajectory points inside }
    \label{fig: time_space}
\end{figure}

This is the most widely used method for measuring macroscopic traffic flow variables when trajectory data of a traffic stream is available \citep{laval2010mechanism, he2019constructing, dahiya2020study, wang2024rectangle, he2025constructing}. This method was originally proposed by \citep{edie1963discussion} and is widely known as Edie's generalized definition of traffic flow variable. Let $n$ denote the total number of vehicles within a given time space region $A$ (as is illustrated in Fig.~\ref {fig: time_space}(a). For the $i^{th}$ vehicles travelling through region $A$, let $x_i$ represent the travel distance and $t_i$ represent the travel time within the time space region. The area of the time-space region is denoted by $|A|$. Using Edie's definitions, the macroscopic traffic variables can be expressed as follows (Eq.~\eqref{eq: macroscopic1}):

\begin{equation}
\begin{aligned}
    k(x, t, A) &= \frac{1}{|A|} \sum_{i=1}^{n} t_i \\
    q(x, t, A) &= \frac{1}{|A|} \sum_{i=1}^{n} x_i \\
    v(x, t, A) &= \frac{q(x, t, A)}{k(x, t, A)} = \sum_{i=1}^{n} \frac{x_i}{t_i}
    \label{eq: macroscopic1}
\end{aligned}
\end{equation}

In this study, $A$ is set to be a parallelogram time space region as shown in Fig. \ref{fig: time_space}(b). The two longer sides of the parallelogram are oriented along the shockwave speed $(w)$ while the other two shorter sides are parallel to the target speed $(v^*)$, as shown in Fig.~\ref{fig: time_space}(b). Prior research \citep{laval2010mechanism, he2019constructing, wang2024rectangle, he2025constructing} has demonstrated that constructing measurement regions whose longer sides align with the direction of shockwave propagation yields more accurate and reliable estimates of macroscopic variables. For the present analysis, the shockwave speed is taken as -18 $km/h$, determined by visually inspecting the time-space diagram of the trajectory dataset. The target speed $(v^*)$ is varied systematically from the jam speed $(0\ km/h)$ to free flow speed $(90 \ km/h)$ using increments of $5 \ km/h$. This ensures that the entire speed range of traffic flow is adequately captured. The dimensions of the parallelogram are chosen as follows: The longer side corresponds to 200 $m$ in the spatial domain, while the longer side spans 5 sec in the temporal domain. These dimensions ensure that, on average, at least 5 vehicle trajectories are included in each parallelogram, helping reduce measurement variability. Importantly, any trajectory passing through this region will be nearly parallel to its shorter side. In this way, we can achieve the instantaneous steady-state conditions, which are essential for constructing a reliable FD.
Under this setup, the travel distance $x_i$ and travel time $t_i$ of the $i^{th}$ vehicle within the parallelogram can be approximated by projecting the shorter sides of the parallelogram onto the spatial and temporal axes, respectively, as illustrated in Fig.~\ref {fig: time_space}(b). Eq.~\eqref{eq: macroscopic1} can be reformulated to accommodate this parallelogram-based region for calculating macroscopic traffic flow variables as follows (Eq.~\eqref{eq: macroscopic}):

\begin{equation}
\begin{aligned}
    k(x, t, A) &= \frac{1}{|A|} \sum_{i=1}^{n} t_i \approx m \frac{t_i}{|A|} \\
    q(x, t, A) &= \frac{1}{|A|} \sum_{i=1}^{n} x_i  \approx m \frac{x_i}{|A|}\\
    v(x, t, A) &= \frac{q(x, t, A)}{k(x, t, A)} = \sum_{i=1}^{n} \frac{x_i}{t_i} = v^*
    \label{eq: macroscopic}
\end{aligned}
\end{equation}
Here, $m$ denotes the total number of vehicles inside the parallelogram, while $x_i$ and $t_i$ represent the travel distance and travel time, respectively. These values remain constant for all vehicles within the region. Therefore, the values of both density and flow are determined exclusively by the number of vehicles present within the defined parallelogram region. The selection of this methodology was inspired by previous studies \citep{he2025constructing}, particularly those focusing on the construction of the FD of traffic flow from large-scale vehicle trajectory data.

\section{Data description and environment setup}
\label{sec: data}
To train the RL agent for longitudinal vehicle control, Leader Vehicle (LV) trajectories are required. Accordingly, this study employs both real-world trajectory data from the NGSIM dataset \citep{usdot_ngsim_2009} and simulated data generated using an Ornstein–Uhlenbeck (OU) process \citep{okhrin2022simulating}, each serving a distinct purpose within the overall analysis framework.

The primary real-world dataset used in this study is the NGSIM trajectory dataset \citep{usdot_ngsim_2009}, which was collected on the eastbound lanes of I-80 in the San Francisco Bay area near Emeryville, CA, on April 13, 2005. The dataset spans a total duration of 45 minutes, divided into three 15-minute time intervals: 4:00 - 4:15 p.m., 5:00 - 5:15 p.m., and  5:15 - 5:30 p.m. These intervals capture different traffic conditions representative of typical freeway operations during peak periods. The dataset provides high-resolution vehicle trajectory information, sampled at 10 Hz, allowing precise computation of positions, speeds, and accelerations. From this data, a total of 1341 car-following events were extracted, in which 70\% (938 pairs) were used for training the RL agent, while the remaining 30\% (403 pairs) were reserved for testing and comparative performance evaluation. The NGSIM data are utilised both to train the RL-based car-following model on realistic driving behavior and to perform a comparative fuel-efficiency analysis against baseline models using real-world trajectories.

\begin{figure}[htpb]
    \centering
    \includegraphics[width=\linewidth]{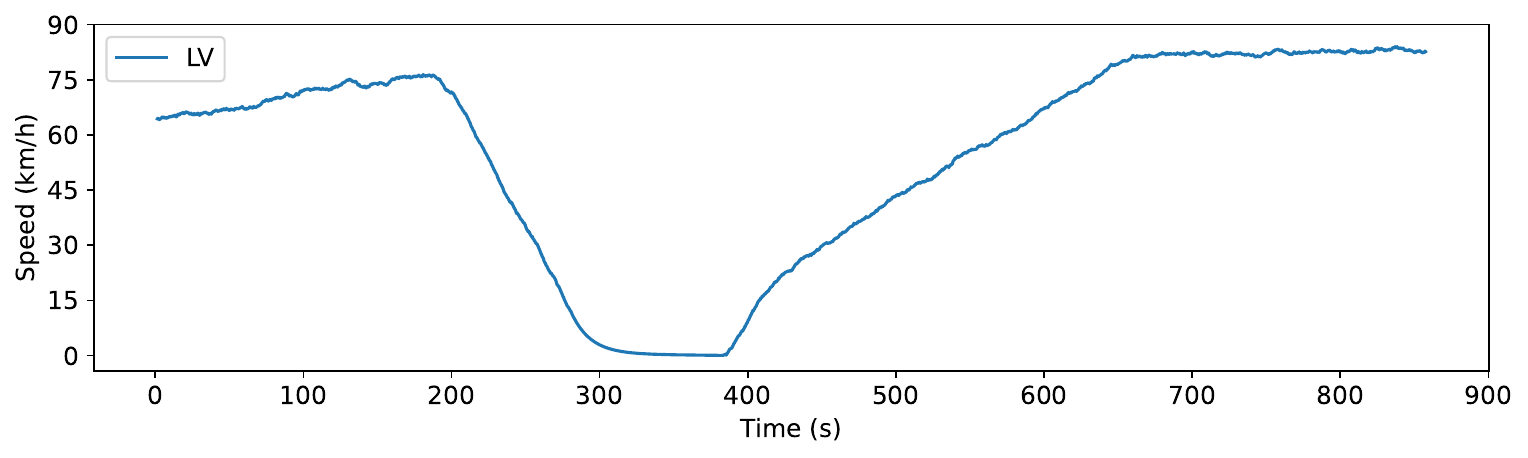}
    \caption{Example leader vehicle trajectories produced by the Ornstein–Uhlenbeck process}
    \label{fig: simdata}
\end{figure}

In addition to real-world data, a simulated dataset is generated using the Ornstein–Uhlenbeck process \citep{okhrin2022simulating} to produce LV trajectories. To obtain the complete profile of the FD, the trajectory of the LV was first generated by systematically controlling its speed from the standstill $(0 \ km/h)$ up to free-flow speed $(v_f = 90 \ km/h)$ at a random interval of $5 \ km/h$ with different phase durations. An example simulated LV trajectory over a 900 sec duration is illustrated in Fig.~\ref{fig: simdata}. This simulated data is employed exclusively for macroscopic traffic flow analysis, particularly for constructing FD. The use of simulated data enables systematic coverage of diverse traffic regimes (e.g., free-flow and congested states), which is difficult to get from real-world datasets.

\section{Results and Interpretation}
\label{sec: test_results}
This section includes the RL training process and overview of the study's findings. The trained RL car-following model is systematically evaluated and compared with that of the IDM, representing human-driven vehicles, within the FD framework to analyse traffic flow behavior under different conditions. We examine the impact of the safe time gap $(T)$ on the FD to assess how driver heterogeneity influences traffic capacity and flow. Mixed traffic scenarios with different penetration rates of RL-based and human-driven (IDM) vehicles are also analysed to evaluate automation effects on overall performance of the traffic stream. Finally, fuel efficiency of the RL model is compared with the IDM to assess potential energy-saving benefits alongside improvements in traffic throughput.

\subsection{Training result}
\label{subsec: training}
The real-world NGSIM dataset (discussed in Section~\ref{sec: data}) has been used to train the RL agent. During training, the RL agent randomly selects a LV trajectory from the training dataset at the start of each episode. Once the episode is terminated, a new LV trajectory is randomly selected for the next episode. This process ensures exposure to diverse traffic scenarios, allowing the agent to learn robust car-following behaviors. The training was conducted over 2,000 episodes, enabling the agent to gradually improve its ability to maintain safety, comfort, and stability of traffic flow across varying traffic conditions.

\begin{figure}[htpb]
    \centering
    \includegraphics[width=\linewidth]{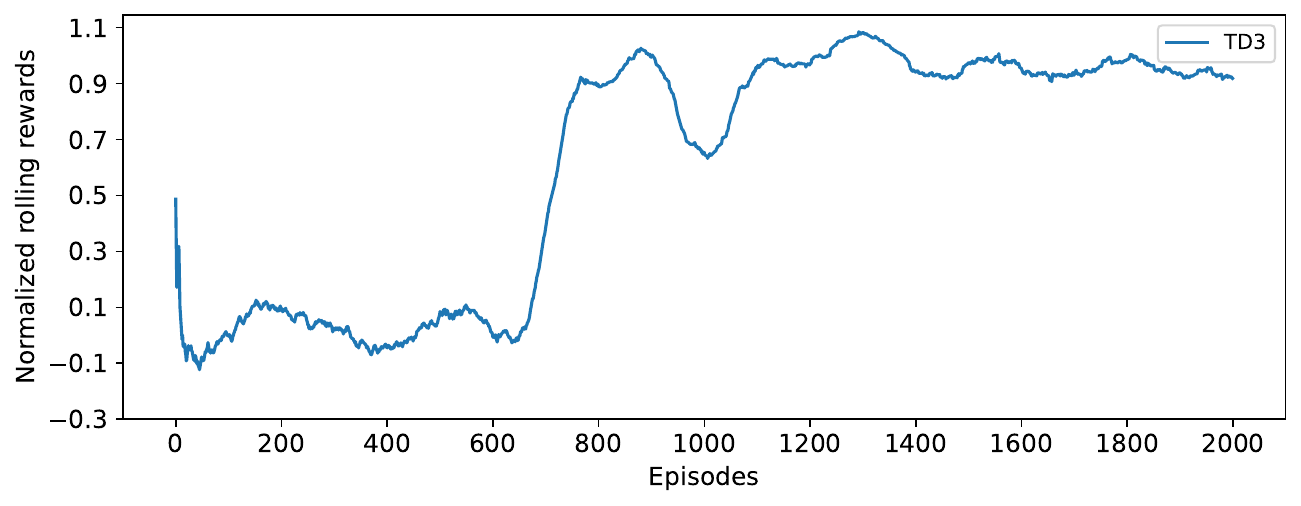}
    \caption{Normalized rolling rewards of the TD3 agent during training}
    \label{fig: training}
\end{figure}

Fig.~\ref{fig: training} illustrates the training performance of the selected TD3 model with the proposed reward function. The y-axis shows the normalized rolling rewards, computed using min-max normalization over a sliding window of 100 steps, while the x-axis indicates the training episode. As shown in Fig.~\ref{fig: training}, during the initial 600 episodes, the normalized rolling reward remains relatively low, reflecting the exploration phase of the model. Around episode 700, a noticeable and rapid increase in the normalized rolling reward is observed, indicating that the agent begins to consistently adopt more effective longitudinal control behaviors. Following this period, the reward gradually stabilises, suggesting that the agent has successfully learned a robust policy. Training was stopped after 2,000 episodes, at which point the learned policy was saved and subsequently used for testing. The results of the testing phase, which evaluate the performance of the trained agent under unseen traffic scenarios, are presented and discussed in the next section.

\subsection{Fundamental diagram under driver heterogeneity}
\label{sec: driver_heterogeneity}
The FD is one of the oldest and most foundational concepts in traffic flow theory for characterising overall traffic flow behavior. The FD typically described the relationship among three macroscopic traffic flow variables: traffic volume (q), density (k), and speed (v). By examining the interactions among these variables under different traffic conditions, the FD provides valuable insights into roadway capacity and the transition between traffic states.

This study employed a macroscopic approach (described in Section~\ref{sec: macroscopic}) to estimate the traffic flow parameters, which requires detailed vehicle trajectory data. To meet this requirement, simulated LV trajectories, as described in Section~\ref{sec: data}, are used as inputs for the first controlled RL-based vehicle, referred to as FV1. The FV1 output trajectory becomes the leader input for the next RL-controlled vehicle (using the same policy), referred to as FV2, and this process continues sequentially, thereby creating a sequence of vehicles. Each follower vehicle in this sequence is governed by the same pre-trained RL agent to ensure consistency in behavior across the simulation. For realism, the initial spacing between the LV and each follower vehicle is randomly assigned within the range of 10 m to 50 m. Moreover, the initial speed and acceleration of each follower vehicle are matched to those of the LV at the start of the simulation. This initialization ensures that the follower vehicles begin from plausible traffic conditions, thereby enhancing the fidelity of the simulated trajectories for FD construction.

\begin{figure}[h]
    \centering
    \includegraphics[width=\linewidth]{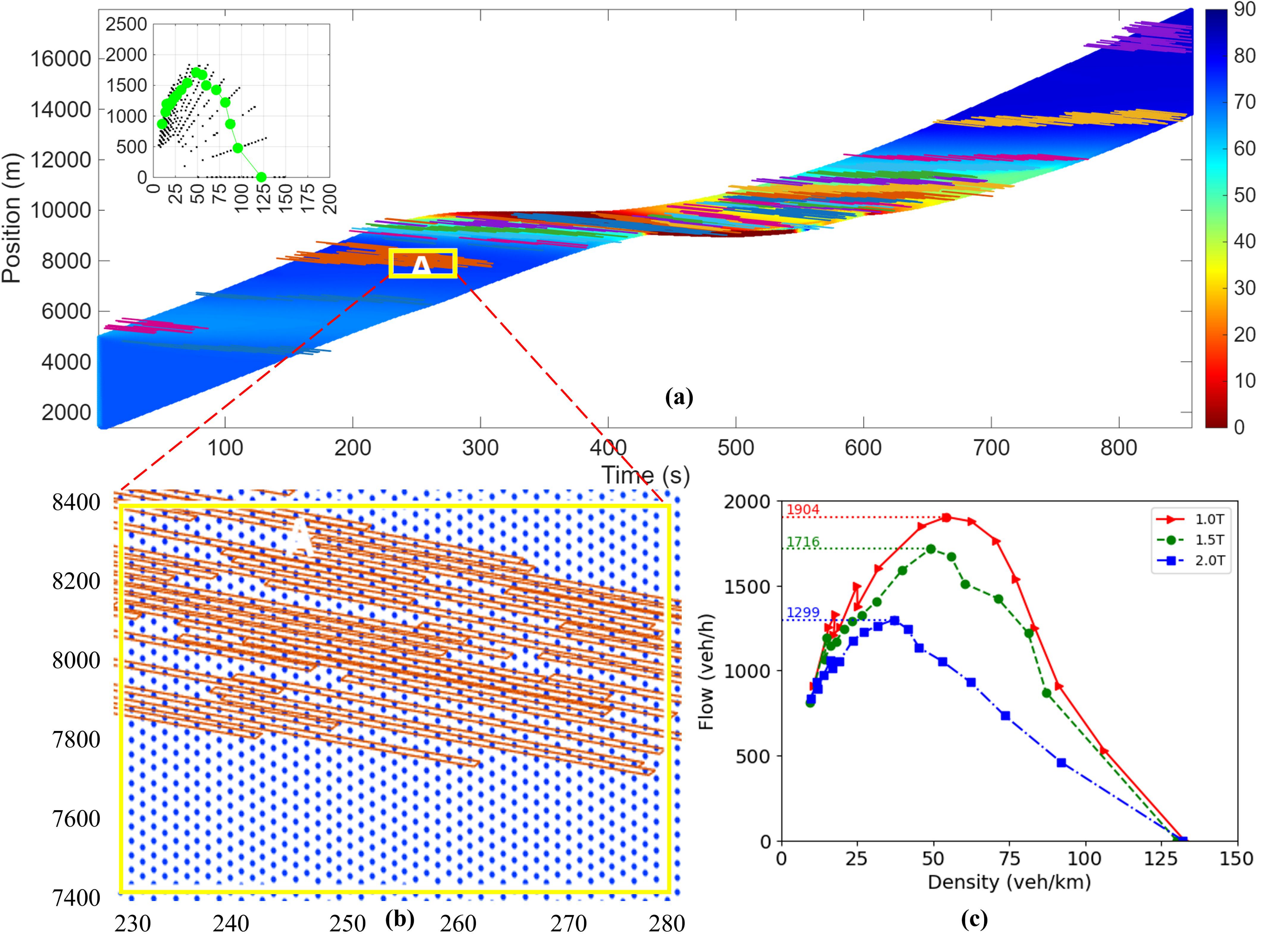}
    \caption{Visualisation of traffic flow analysis: (a) time–space trajectory data with generated parallelogram, (b) magnified section of region A, and (c) empirical fundamental diagrams under varying levels of driver heterogeneity}
    \label{fig: density}
\end{figure}

Within this framework, driver heterogeneity is incorporated by adjusting the desired time gap parameters $(T)$ in Eq.~\eqref{eq: space_headway}. A lower value of $T$ encourages the RL agent to follow the leading vehicle more closely, representing a more aggressive driving style, whereas a higher $T$ results in larger spacings that mimic more conservative behavior. To explore the impact of this heterogeneity on traffic dynamics, three representative time headway values are considered: $T = 1$ sec, $T = 1.5$ sec, and $T = 2$ sec. For each case, the pre-trained RL model is executed to generate the corresponding sequence of vehicle trajectories. This study has taken 100 trajectories to ensure sufficient coverage of the entire speed range, from jam conditions to free-flow, as illustrated in Fig.~\ref{fig: density}(a). The magnified view of region $A$ is provided in Fig.~\ref{fig: density}(b), which clearly demonstrates that the parallelogram regions are constructed only in areas that satisfy the methodological conditions discussed in Section~\ref {sec: macroscopic}. The vehicle trajectories that intersect these parallelogram regions are subsequently used to compute aggregate traffic variables, including density, flow, and speed. The resulting FD, corresponding to the different levels of driver heterogeneity, are presented in  Fig.~\ref{fig: density}(c). Preliminary tests indicated that increasing the number of trajectories beyond 100 does not produce appreciable changes in the results, but significantly increases computational time. Therefore, 100 trajectories were deemed sufficient for reliable analysis.

As shown in Fig.~\ref{fig: density}(c), the red solid, green dashed, and blue dash-dotted curves represent the average flow-density relationships corresponding to time headway of $T = 1$ sec, $T = 1.5$ sec, and $T = 2.0$ sec, respectively. These curves are obtained from the DRL-based car-following simulations and illustrate how variations in driver time headway influence macroscopic traffic behavior.
The results demonstrate that the traditional FD structure remains applicable for describing the driver heterogeneity introduced through different safe time gaps in the RL-based control framework. This indicates that, despite the use of learning-based automated driving policies, the aggregate traffic dynamics are consistent with well-established traffic flow theory. A clear shift in traffic performance is observed in the safe time gap $(T)$ increases. Specifically, a larger safe time gap $(T)$ reduces overall road efficiency, as reflected in lower traffic flow capacity. This reduction occurs because vehicles operating with a higher time gap maintain larger longitudinal gaps, which decreases the number of vehicles that can occupy a given road segment at the same density. While this behavior improves safety and driving comfort at the microscopic level, it results in lower throughput at the macroscopic level.
Overall, these findings highlight the inherent trade-off between safety-oriented driving behavior and traffic efficiency. The DRL-based car-following model effectively captures this trade-off, producing realistic macroscopic traffic patterns while reflecting the impact of driver heterogeneity through varying time gaps.

\subsection{Fundamental diagram under varying penetration rates of driver heterogeneity}
\label{sec: T_penetration}

Apart from evaluating the impact of individual time gap values $(T)$ (driver heterogeneity) on the FD, this section focuses on the impact of driver heterogeneity with different proportions of aggressive and safe drivers in the traffic stream. This has been represented by different proportions of RL-agents with different safe time gaps ($T$)  parameter values in the traffic stream. This analysis is particularly important from a practical perspective because, in real traffic streams, drivers maintain different time gaps, reflecting their risk-taking attitudes. Therefore, in a traffic stream, a range of time gaps is typically observed simultaneously. To capture a more realistic traffic composition, a mixed time-gap scenario has been considered. In this scenario, 60\% of the vehicles were assigned to a one-time gap value, while the remaining 40\% of vehicles were distributed equally among the other two gap values. For example, 60\% of vehicles had $T = 1$ sec, while 20\% each had $T = 1.5$ sec, and $T = 2$ sec, with similar distributions for other combinations. Additionally, an equal-percentage distribution across all considered time gaps was also tested. In all cases, time gaps in assignments were randomised within the defined proportions.

Table~\ref{table: T_penetration} summarises the estimated FD parameters, including optimal density and capacity for different levels of driver heterogeneity, represented through mixed safe time-gap distributions. Each row corresponds to a specific combination of time-gap values and their associated penetration rates within the traffic stream. Each combination reflects a realistic traffic composition in which the driver follows a different headway preference rather than a single uniform value.

The results indicate that macroscopic traffic performance is strongly influenced by the proportion of vehicles operating with shorter time gaps. When a larger share of vehicles adopts a smaller time gap, such as when 60\% use $T = 1$ sec, the optimal traffic density and corresponding capacity are highest, at 51 $veh/km$ and 1773 $veh/h$, respectively. In contrast, as the percentage distribution shifted towards a large time gap, both the optimal density and capacity decreased. For instance, when 60\% of vehicles follow a more conservative time gap of $T = 2$ sec, the optimal density reduces to 43 $veh/km$, and the capacity reduces to 1508 $veh/h$. The equal-distribution scenario (33\% each time gap) yields intermediate performance levels, lying between the two extremes.

These findings are consistent with the observations presented in Section~\ref{sec: driver_heterogeneity}. The RL-based car-following model reproduces the classical FD structure at the macroscopic level, despite incorporating learning-based vehicle control. Furthermore, the observed reduction in capacity with increasing time gaps aligns well with previous studies on mixed traffic involving automated vehicles, CACC, and human-driven vehicles \citep{li2023fundamental, wang2023fundamental, jiang2024fundamental}. 
Overall, these results highlight that increased driver heterogeneity and a higher prevalence of conservative time gaps reduce road capacity, whereas a greater share of shorter time gaps leads to more efficient utilisation of available road space.

\begin{table}[htpb]
    \caption{Estimated FD parameters corresponding to different penetration rates of driver heterogeneity}
    \label{table: T_penetration}
    \centering
    \renewcommand{\arraystretch}{1.1} 
    \begin{tabular*}{1\textwidth}{@{\extracolsep{\fill}} l c c }
        \toprule
        \makecell{Penetration rate of driver \\  heterogeneity (\%)} & \makecell{ Optimal traffic density \\ (veh/km)} & \makecell{Traffic capacity \\(veh/h)} \\
        \midrule
        $\left[1.0T \ (60\%), 1.5T \ (20\%), 2T \ (20\%) \right]$ & 51 & 1773 \\
        $\left[1.0T \ (20\%), 1.5T \ (60\%), 2T \ (20\%) \right]$ & 49 & 1683 \\
        $\left[1.0T \ (34\%), 1.5T \ (33\%), 2T \ (33\%) \right]$ & 47 &  1632 \\
        $\left[1.0T \ (20\%), 1.5T \ (20\%), 2T \ (60\%) \right]$ & 43 & 1508 \\
        \bottomrule
    \end{tabular*}
\end{table}

\subsection{Fundamental diagram of mixed traffic flow under varying penetration rates}
\label{sec: RL_penetration}

In addition to evaluating the impact of driver heterogeneity and their penetration rates on FD, it is also equally important to investigate the effect of the mixed traffic flow under varying penetration rates of RL automated vehicles. In this context, mixed traffic flow refers to a scenario in which both human-driven vehicles (represented using classical car-following models, such as the IDM) and RL-based autonomous agents coexist on the same roadway. This reflects a realistic transitional phase of traffic systems, where human-driven (IDM) vehicles and automated vehicles (RL) will share the road for an extended period before full automation is achieved. Understanding the characteristics of mixed traffic flow is therefore critical for the accurate analysis, modelling, and simulation of future traffic systems as the penetration rate of automated vehicles gradually increases.

In this study, the RL-based model is regarded as a fully automated vehicle, while the IDM is employed to capture the behavior of human-driven vehicles. To explore the impact of automation on overall traffic performance, the penetration rate of RL-based automated vehicles within the mixed traffic stream was systematically varied across five levels: 0\%, 25\%, 50\%, 75\%, and 100\%, with the time headway setting $T = 1.5$ sec. These scenarios enable a direct assessment of how gradual increases in automation impact the fundamental characteristics of traffic flow. The key parameters of the FD, including the optimal traffic density and the corresponding traffic capacity, are summarized in Table~\ref{table: RL_penetration}.
As the penetration of RL-based vehicles increases, both the optimal density and corresponding traffic flow consistently rise. This trend indicates an improvement in road utilisation due to more efficient vehicle interactions. The transition from fully human-driven vehicles (0\% RL) to fully RL-based automated vehicles (100\% RL) increases road efficiency. The estimated road capacity increases from 1596 veh/h to 1716 veh/h, resulting in an improvement of approximately 7.52\%. However, the increase in capacity is relatively small at partial penetration levels of RL-based vehicles. These findings are consistent with previous studies on mixed traffic flow that investigate the effects of automated vehicle technologies, such as ACC and CACC. For instance, \citep{li2023fundamental} shows that higher penetration rates of ACC and CACC vehicles increase traffic capacity and critical density, whereas low to moderate penetration rates yield only modest improvements. Similarly, \citep{wang2023fundamental, jiang2024fundamental} demonstrates that increasing CACC penetration gradually stabilises traffic flow and enhances capacity, but substantial macroscopic benefits appear only at high penetration levels. The limited improvements at partial penetration are due to the continued influence of human-driven vehicles, whose variable behavior constrains the stabilizing effect of automated vehicles. These results align with our observation that partial adoption of RL‑based automated vehicles produces limited capacity gains until penetration approaches full saturation.

\begin{table}[htpb]
    \caption{Estimated FD parameters for different penetration rates of RL-based vehicles with a fixed safe time gap of $T=1.5$ sec}
    \label{table: RL_penetration}
    \centering
    \renewcommand{\arraystretch}{1.1} 
    \begin{tabular*}{1\textwidth}{@{\extracolsep{\fill}} c c c c }
        \toprule
         \makecell{Penetration rate\\of RL (\%)} & \makecell{ Optimal traffic  \\  density (veh/km)} & \makecell{Traffic capacity \\(veh/h)} & \makecell{Increase in  \\ capacity (\%)} \\
        \midrule
        \textbf{100} & \textbf{50} & \textbf{1716}  & \textbf{7.52} \\
        75 & 47  & 1632  & 2.25 \\
        50 & 46 & 1614  & 1.13 \\
        25 & 45 &  1609  & 0.82 \\
        0 & 45 & 1596  & 0 \\
        \bottomrule
    \end{tabular*}
\end{table}

\subsection{Comparative fuel efficiency analysis}
\label{sec: fuel}
\begin{figure}[h]
    \centering
    \includegraphics[width=0.98\linewidth]{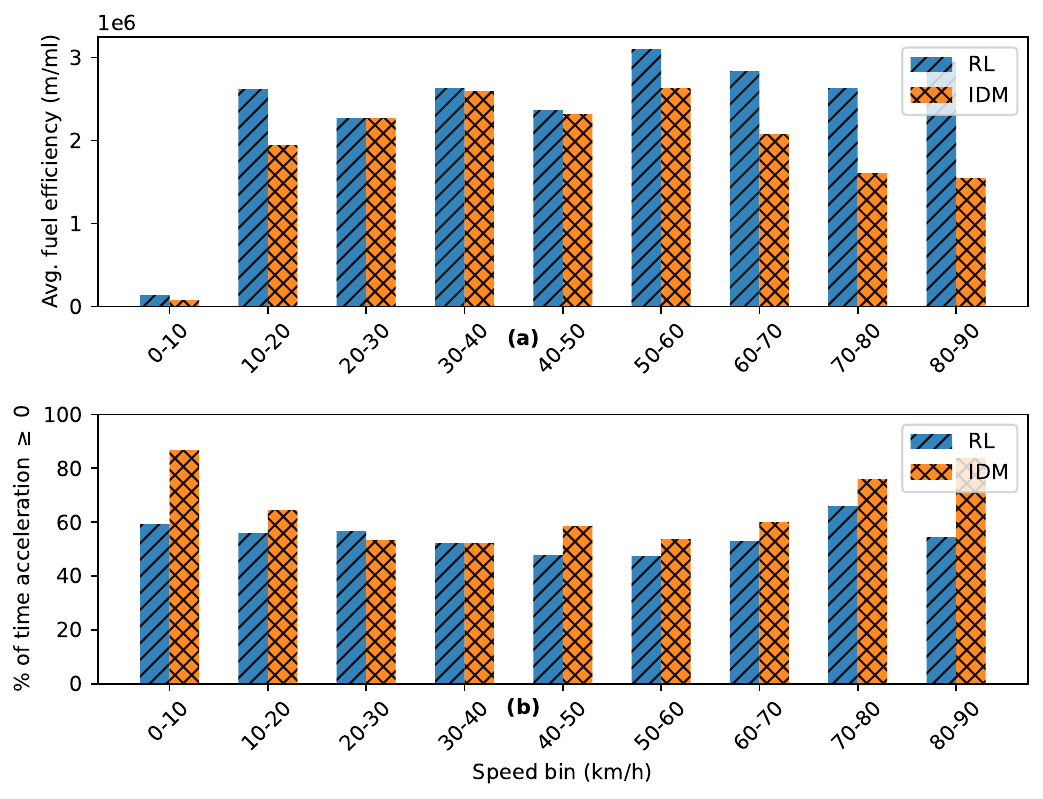}
    \caption{Fuel efficiency performance of RL and IDM models: (a) average fuel efficiency, and (b) \% of time with positive acceleration across speed ranges}
    \label{fig: fuel}
\end{figure}

In addition to evaluating the RL model's performance for traffic-flow characteristics across various scenarios, we now extend our analysis to explore its implications on energy performance. This section presents a detailed analysis of the fuel efficiency achieved by the trained RL model, and the results are systematically compared with those obtained from the IDM as shown in Fig.~\ref{fig: fuel}(a) and (b) to assess potential improvements in energy consumption. 

For the fuel consumption analysis, vehicle speeds were first grouped into 10 $km/h$ intervals (0-10 $km/h$, 10-20 $km/h$, ..., 80-90 $km/h$). For each bin speed, the average fuel efficiency was calculated as the fuel consumed per unit distance travelled. The analysis was performed exclusively on the NGSIM test dataset (containing 403 LV trajectories, as described in Section~\ref{sec: data}) with a fixed safe time gap of $T = 1.5$ sec. The NGSIM dataset was chosen because it captures real-world driving behavior, including natural variations in speed, acceleration, and vehicle interactions, whereas simulated data may include unrealistic driving scenarios that do not accurately reflect fuel consumption patterns in real traffic conditions. 

The fuel efficiency results are compared with the IDM rather than directly with human-driven vehicles. This is because most existing benchmark datasets, such as NGSIM \citep{usdot_ngsim_2009} and HighD \citep{highDdataset}, are collected using high-altitude cameras or drones. Therefore, vehicle position information from images has been used to estimate vehicle acceleration and deceleration. The inherent noise in the image-based position data becomes more pronounced in the acceleration and deceleration data, thereby not representing the actual human acceleration-deceleration profiles. Previous studies, such as \citep{coifman2017critical, wu2024traffic}, have applied Gaussian filtering to reduce noise, yet the resulting acceleration profiles still differ from those obtained directly from vehicles or models. Since fuel consumption (Eq.\eqref{eq: fuel_rate}) is directly proportional to speed and acceleration, even small fluctuations can lead to overestimation of fuel usage. By comparing with IDM, which provides a well-established baseline for human-based car-following behavior, we can more reliably assess the RL-based model's fuel efficiency under controlled and realistic assumptions. Future work will extend this comparison to fuel data directly measured from vehicle engines for further validation.

The resulting average fuel efficiency across different speed ranges is presented in Fig.~\ref{fig: fuel}(a), where the blue line and orange dot line denote the RL model and IDM, respectively. The RL-based car-following model improves average fuel efficiency by approximately 28.98\% at higher speeds (above 50 $km/h$), while at lower speeds (below 50 $km/h$) the improvement is around 1.86\%. The observed fuel efficiency difference can be attributed to the variation in acceleration behavior between the two models, as illustrated in Fig.~\ref{fig: fuel}(b). This figure shows that the percentage of time acceleration is greater than or equal to zero across different speed bins. The IDM exhibits a higher proportion of positive acceleration compared to the RL-based model, especially at higher speeds. Since positive acceleration directly increases the engine's power demand, more frequent acceleration events under IDM lead to higher fuel consumption. In contrast, during deceleration, propulsion power is effectively reduced to zero (Eq.~\eqref{eq: fuel_power}), and fuel consumption is limited to the idle engine usage (Eq.~\eqref{eq: fuel_rate}).

Overall, these results demonstrate that the RL-based car-following strategy not only reproduces realistic traffic flow behavior but also substantially improves energy efficiency compared to a conventional car-following model like the IDM, particularly at higher speeds.
Together with the previous findings, the use of a shorter time headway under fully RL-controlled traffic conditions can improve both traffic flow capacity and fuel efficiency without compromising the traffic safety or overall performance. These results indicate that the interaction between driver heterogeneity and the penetration rate of automated vehicles plays an important role in determining overall traffic performance and fuel efficiency in mixed-traffic environments.

\section{Conclusion}
\label{sec: conclusion}
This study investigated the macroscopic traffic flow characteristics of mixed traffic consisting of DRL-based automated vehicles and human-driven vehicles (IDM). A systematic analysis has been done to estimate the macroscopic traffic variables such as flow and density using FD under varying (i) driver heterogeneity, (ii) different penetration rate of driver heterogeneity, and (iii) different percentage of RL-based automated vehicles. Additionally, a comparative fuel-efficiency analysis was conducted between the RL-based automated and the human-driven (IDM) vehicles. 

To generate the required microscopic car-following trajectories, a TD3 algorithm has been utilised to develop an adaptive and robust car-following model using the given reward function. The RL agent was trained on real-world, naturalistic driving data from the NGSIM datasets, enabling it to learn realistic driving behavior across varying traffic conditions. For comparative analysis, car-following trajectories were generated using both the proposed RL-based model and the conventional IDM. These microscopic trajectories were aggregated into macroscopic traffic measures using Edie’s method, which estimates traffic flow variables, including speed, density, and flow. Based on the resulting FD, the optimal traffic density and the corresponding flow, representing road capacity, were identified under different traffic scenarios and systematically compared with those obtained using the traditional IDM.

The results show that traffic performance is strongly influenced by both safe time gap (driver heterogeneity) and its distribution across the traffic stream. A reduction in the safe time gap increases road capacity under the RL-based automated vehicle framework. When the penetration rate of different time gaps is varied, traffic performance is primarily governed by the proportion of vehicles operating with shorter time gaps. For instance, a scenario with 60\% of vehicles following a shorter time gap of $T=1$ sec achieves a road capacity of 1773 veh/h, whereas the capacity decreases to 1508 veh/h when 60\% of vehicles adopt a more conservative time gap of $T=2$ sec. Furthermore, increasing the penetration of RL-based automated vehicles consistently improves traffic efficiency. The transition from fully human-driven traffic (0\% RL) to fully RL-controlled traffic (100\% RL) resulted in an approximate 7.52\% increase in road capacity. This improvement can be attributed to smoother car-following behavior and reduced unnecessary acceleration and deceleration under the RL-based control strategy. Finally, fuel efficiency results further support the advantages of the proposed approach. Using real-world NGSIM trajectory data, the RL-based automated vehicle demonstrated an average fuel efficiency improvement of approximately 28.98\% at higher speeds (above 50 $km/h$), and about 1.86\% at lower speeds (below 50 $km/h$) compared to the IDM. This gain is primarily due to reduced positive acceleration events and smoother speed profiles, particularly at higher speeds.

Overall, the findings confirm that an RL-based car-following strategy, combined with a moderate time headway of $T=1.5$ sec and high penetration of automated vehicles, can enhance road capacity and fuel efficiency without compromising traffic safety. These results highlight the potential of RL-based control strategies to support more efficient and sustainable traffic operations in future mixed traffic environments.

In terms of limitations, this study relies on a single LV trajectory from simulated data to generate the subsequent vehicle trajectory for estimating road capacity and fuel efficiency, which is then compared with the traditional IDM. As a result, the interaction effects arising from different LV vehicles and vehicle types are not considered. In future work, the framework can be extended to incorporate interactions with multiple lead vehicles and various vehicle types and evaluate fuel efficiency using directly measured fuel consumption data from vehicle engines, thereby providing a more accurate representation of real-world traffic conditions and fuel efficiency.

\section*{CRediT authorship contribution statement}
\textbf{Pankaj Kumar:} Conceptualisation, Formal analysis, Methodology, Investigation, Writing.
\textbf{Pranamesh Chakraborty:} Conceptualisation, Methodology, Investigation, Supervision, Writing.  
\textbf{Subrahmanya Swamy Peruru:} Conceptualisation, Methodology, Investigation, Supervision, Writing.  

\section*{Declaration of competing interest}
The authors confirm that they have no financial or personal conflicts of interest that could have influenced the findings of this study.

\bibliography{bibliography}
\end{document}